# *FdConfig*: A Constraint-Based Interactive Product Configurator

Denny Schneeweiss and Petra Hofstedt

Brandenburg University of Technology, Cottbus
`schneden@tu-cottbus.de` and `hofstedt@informatik.tu-cottbus.de`

**Abstract.** We present a constraint-based approach to interactive product configuration. Our configurator tool *FdConfig* is based on feature models for the representation of the product domain. Such models can be directly mapped into constraint satisfaction problems and dealt with by appropriate constraint solvers. During the interactive configuration process the user generates new constraints as a result of his configuration decisions and even may retract constraints posted earlier. We discuss the configuration process, explain the underlying techniques and show optimizations.

## 1 Introduction

Product lines for mass customization [22] allow to fulfill the needs and requirements of the individual consumer while keeping the production costs low. They enhance extensibility and maintainance by re-using the common core of the set of all products.

*Product configuration* describes the process of specifying a product according to user-specific needs based on the description of all possible (valid) products (the search space). When done *interactively*, the user specifies the features of the product step-by-step according to his requirements, thus, gradually shrinking the search space of the configuration problem. This interactive configuration process is supported by a software tool, the *configurator*.

In this paper, we present an approach on interactive product configuration based on constraint programming techniques. Building on constraints enables us to equip our interactive product configurator *FdConfig* with functionality and expressiveness exceeding traditional approaches but at the cost of performance penalty which must be dealt with in turn.

The paper is structured as follows: In Sect. 2 we briefly review the area of interactive configuration methods and discuss related work. Section 3 introduces important notions from the constraint paradigm as needed for the discussion of our approach. We present the constraint-based interactive product configurator *FdConfig* in Sect. 4. There, we introduce *FdFeatures*, a language for the definition of feature models, it's transformation into constraint problems, and the configuration process using *FdConfig*. Furthermore, we discuss optimizations and improvements by analyses and multithreading. Section 5 draws a conclusion and points out directions of future research.

## 2 Interactive Configuration Methods

An interactive product configurator is a tool which allows the user to specify a product according to his specific needs based on the common core of the set of all products of a product line. This process can be done interactively, i.e. in a step-wise fashion, thus gradually shrinking the search space of the configuration problem.

For the sake of applicability and user-friendliness, a configurator requires a number of properties like backtrack-freeness, completeness, order-independent retraction of decisions, short response times and others. These strongly depend on the method[1] underlying the configurator system. Cost optimization and arithmetic constraints are a desired functionality too, but these are seldom supported or only provided in a very restricted form.

---

[1] For a discussion of the solutions methods see below.

While *completeness* ensures that no solutions are lost, *backtrack-freeness* [7,20] guarantees that the configurator only offers decision alternatives for which solutions remain. Thus, the user can always generate a valid solution from the current configuration state and does not need to unwind a decision (i.e. he does not need to backtrack). The *Calculate Valid Domains (CVD) function* [7] of a configurator realizes this latter property.

*Feature models* are particularly used in the context of software product line engineering to support the reuse when building software-intensive products. However, they are of course applicable to many other product line domains. They stem from the *feature oriented domain analysis methodology* (FODA) [14].

A feature model describes a product domain by a combination of features, i.e. specific aspects of the product which the user can configure by instantiation and further constraints. A product line is given by the set of possible combinations of feature alternatives.

The semantics of feature models is typically mapped to propositional logics [11] and can accordingly be mapped onto a restricted class of constraint satisfaction problems (cf. Sect. 3), namely constraints of the Boolean domain. While many approaches in the literature (e.g. [7,8,12]) only consider constraints of the Boolean domain (including equality constraints), Benavides et.al. [1] discuss the realization of arithmetic computations and cost optimization in a feature model (by so-called "extra-functional features") which can be represented by general constraint problems.

*Solution techniques* applied to the interactive configuration problem have been compared by Hadzic et.al. [7,8] and Benavides et.al. [2]. They mainly distinguish approaches based on propositional logic on the one hand and on constraint programming on the other hand.

When using *propositional logic* based approaches, configuration problems are restricted to logic connectives and equality constraints (see e.g. [7,21]). Arithmetic expressions are excluded because of the underlying solution methods. These approaches perform in two steps. First, the feature model is translated into a propositional formula. In the second step the formula is solved (satisfiability checking, computation of solutions) by appropriate solvers, in particular *SAT solvers* (as in [12]) and *BDD-based solvers* (see e.g. [8,20]). BDD-based solvers translate the propositional formula into a compact representation, the BDD (binary decision diagram). While many operations on BDDs can be implemented efficiently, the structure of the BDD is crucial as a bad variable ordering may result in exponential size and, thus, in memory blow up. Therefore the compilation of the BDD is done in an offline phase, so a suitable variable ordering can be found and the BDD's size becomes reasonably small.

Feature models can be naturally mapped into *constraint systems*, in particular into CSPs. There are some approaches [1,21] using this correspondence to deal with interactive configuration problems. These typically work as follows: The feature model is translated into a constraint satisfaction problem (CSP, see Definition 1 below) and afterwards analysed by a CSP solver. Using this approach, no pre-compilation is necessary. In general it is possible to use predicate logic expressions and arithmetic in the feature definitions, even if this is not realized in the above mentioned approaches.

Transformations of feature models into programs of CLP languages (i.e. Prolog systems with constraints) have been shown recently in [15,17]. However, beside the transformation target beeing different from ours, these approaches do not focus on using these methods for interactive configuration.

Since our *FdConfig* tool aims primarily at the software engineering community as the main users of feature models, we decided in favour of a Java-implementation, which would make later integration with common software development infrastructure like *Eclipse* more easy.

Benavides et.al. [2] elaborately compare the approaches sketched above, particularly with respect to *performance* and *expressiveness or supported operations, resp.* They point out that CSP-based approaches, in contrast to others, can allow extra functional features [1,14] and, in addition, arithmetic and optimization. Furthermore, they state that "the available results suggest" that constraint-based and propositional logic-based approaches "provide similar performance", except for the BDD-approach which "seems to be an exception as it provides much faster execution times", but with the major drawback of BDDs having worst-case exponential size.

Extended feature models with numerical attributes, arithmetic, and optimization are denominated as an important challenge in interactive configuration by Benavides et.al. [2]. Our approach aims at this challenge. The main idea is to follow the constraint-based approach while using the combination of different constraint methods and concurrency to deal with the computational cost. At this, a major challange is to support the user when making and withdrawing decisions in an interactive process.

## 3 Constraint Programming

Feature models can directly be mapped into corresponding constraint problems. We will discuss this approach more detailed in Sect. 4.1 but introduce the necessary notions from the constraint paradigm here.

*Constraints* are predicate logic formulae which express relations between the elements or objects of the problem. They are classified into *constraint domains* (see [9,18]), e.g. linear constraints, Boolean constraints and finite domain constraints. This partitioning is due to the different applicable constraint solution algorithms implemented in so-called constraint solvers (see below).

Considering feature models as constraint problems, the domains of the involved variables are a priori finite.[2] Thus, we consider a particular class of constraints: *finite domain constraints.* Finite domain constraint problems are given by means of constraint satisfaction problems.

**Definition 1 (CSP).** *A* Constraint Satisfaction Problem (CSP) *is a triple $P = (X, D, C)$, where $X = \{x_1, \ldots, x_n\}$ is a finite set of variables, $D = (D_1, \ldots, D_n)$ is an n-tuple of their respective finite domains, and $C$ is a conjunction of constraints over $X$.*

**Definition 2 (solution).** *A* solution *of a CSP $P$ is a valuation $\varsigma : X \to \bigcup_{i \in \{1,\ldots,n\}} D_i$ with $\varsigma(x_i) \in D_i$ which satisfies the constraint conjunction $C$.*

A CSP can have one solution or a number of solutions, or it can be unsatisfiable. Optimization functions may also be given which specify optimal solutions.

*Example 1.* Consider a CSP $P = (X, D, C)$ with the set $X = \{Cost, Color, Band\}$ of variables and their respective domains $D = (D_{Cost}, D_{Color}, D_{Band})$ with $D_{Color} = \{Red, Gold, Black, Blue\}$, $D_{Cost} = \{0, ..., 1400\}$, and $D_{Band} = \{700, 800, 1000\}$.

$C = (Band = 700 \to Color = Blue) \land (Cost = Band + 500)$ is a conjunction of constraints over the variables from $X$.

Solutions of the CSP $P$ are e.g. $\varsigma_1$ with $\varsigma_1(Cost) = 1200$, $\varsigma_1(Color) = Blue$, and $\varsigma_1(Band) = 700$ which is also denoted by $\varsigma_1 = \{Cost/1200, Color/Blue, Band/700\}$ and $\varsigma_2 = \{Cost/1300, Color/Red, Band/800\}$.

*Constraint solvers* are sets or libraries of tests and operations on constraints, which are able to check the satisfiability of constraints and to compute solutions and implications of constraints.

CSPs are typically solved by narrowing the variable's domains using search nested with consistency techniques (e.g. node, arc, and path consistency). Given a CSP, in the first step consistency techniques are applied. Such consistency checking algorithms work on $n$-ary constraints and try to remove values from the variables domains which cannot be elements of solutions. Afterwards, search is initiated, e.g. using backtracking, where we assign domain values to variables and perform consistency techniques to narrow the other variable's domains again. This search process is controlled by heuristics on variable and value ordering (for the complete process, see [18]).

---

[2] An extension to infinite domains would be possible, in general.

There are some finite domain constraint solver libraries available, for example the JAVA-libraries CHOCO [3] as well as JACOP [10] and the C++-library GECODE [6]. We decided in favor of the freely available CHOCO library which is under continuous development.

Additionally, we need the notions of global consistency and of valid domains.

**Definition 3 (global consistency, see [18]).** *A CSP is i-consistent iff given any consistent instantiation of $i-1$ variables, there exists an instantiation of any ith variable such that the i values taken together satisfy all of the constraints among the i variables. A CSP $P = (X, D, C)$ is globally consistent, if it is i-consistent for every $i$, $1 \leq i \leq n$, where n is the number of variables of C.*

**Definition 4 (valid domains).** *Given a CSP P, the* valid domains *of P is an n-tuple $D_{vd} = (D_{vd,1}, \ldots, D_{vd,n})$ such that each $D_{vd,i} \subseteq D_i$ contains exactly the values which are elements of solutions of P.*

So, if a CSP is gobally consistent, then its domains are valid domains.

*Example 2.* (continuation of Example 1) The valid domains of the CSP $P$ is $D_{vd} = (\{1200, 1300\}, D_{Color}, \{700, 800\})$.

## 4 The Interactive Configurator *FdConfig*

Our approach on interactive configuration consists of two phases: In the first phase a feature model is analysed and then transformed into a CSP and passed to the CHOCO solver. Afterwards the interactive configuration phase follows.

Figure 1 illustrates the analysis and transformation phase. *FdConfig* uses *FdFeatures* files as input. *FdFeatures* is a textual domain specific language for extended feature models which supports integer feature attributes and arithmetic constraints. An *FdFeatures* parser reads the input-file and creates the feature model which is transformed into a CHOCO CSP. Section 4.1 describes the language *FdFeatures* and the transformations in greater detail. Additionally, a quick pre-calculation of the variable's domains is performed. It generates redundant constraints which, nevertheless, help to improve the solver's performace.[3] This *domain analysis* is covered in Sect. 4.2.

In the second phase, the generated CSP is passed to the CHOCO solver which reads the model and creates an internal representation from it: the solver model. Then the solver is started to perform an initial calculation of consequence decisions that yield from the constraints in the *FdFeatures* model. Afterwards, the user can start with the interactive configuration. The implementation of this process is explained in Sect. 4.3. Section 4.4 describes the reduction of response times by using multithreading.

### 4.1 *FdFeatures* Models and CSPs

*FdConfig* provides *FdFeatures* as a language for the definition of feature models based on the approach of [5]. *FdFeatures* borrows from the TEXTUAL VARIABILITY LANGUAGE (TVL, [4]) but was adapted for our needs (e.g. including support for the realization of the user interface, certain detailed language elements and syntactic sugar). *FdFeatures* has been implemented using XTEXT [23].

An *FdFeatures feature model* in general has a tree structure, i.e. there is a distinguished root feature which stands for the item to be configured, but apart from this behaves like any other feature. The model may have additional constraints between (sub-)features and their attributes which, in fact, makes the tree a general graph. Nevertheless, the tree structure is dominant.

---
[3] In constraint programming, the generation of redundant constraints from a given constraint problem is a frequently used method which helps to speed up the solver (see [18], Sect. 12.4.5). Note that the elimination of verification-irrelevant features and constraints (i.e. "redundant relationships", [24]) from feature models with the aim of reducing the problem size is a different concept.

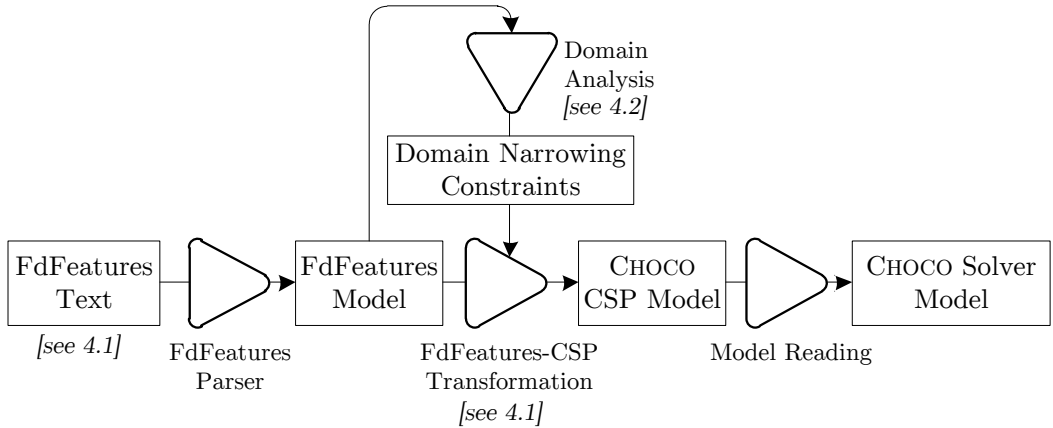

**Fig. 1.** Transformations performed before the user can start configuring

A feature may consist of sub-features and attributes (both in general optional), where, following the approach of [5], the sub-features can be organized in *feature groups*. A feature group allows to describe whether one, some, or all of the sub-features must be included in the configured product.

With similar effects, *features* can be specified to be **mandatory** or **optional**. Furthermore, features may exclude or require other features.

*Example 3.* Consider the cut-out of a feature model of events organized by an event agency in Listing 4.1.[4] For an event (the root feature) we can optionally order a band and a stage, but we must order a carpet (e.g. for a film premiere or a wedding) and colored balloons. These are all modeled as sub-features (which are not organized in a feature group here). Ordering a band makes a stage necessary, expressed by the **requires**-statement in Line 11.

*FdFeatures* supports three kinds of feature *attributes:* integers, enumerations, and Boolean values.

*Example 4.* (continuation of Example 3) The feature *Carpet* is determined by several attributes, including an enumeration attribute *Color*, whose domain elements must be given explicitly and a Boolean attribute *SlipResistance*.[5]

*Length* and *Breadth* are integer attributes. While *Breadth* is specified by an interval, *Length* is unbounded. As we can see by the attributes of *ColoredBalloons*, the domain of an integer attribute can also be specified by a finite set (*Amount*, Line 20) or even by an arithmetic formula (*Cost*, Line 25). The definition of Boolean attributes is also possible using Boolean expressions (but is left out in our example).

The domain definition of *PriceReduction.Coupon* (Lines 22, 23) uses *Guards* to define the attribute domain depending on the configuration state (**ifIn** and **ifOut** correspond to selected

---

[4] The description of certain features and attributes, which are not necessary for the understanding of this example and the concepts behind, is left out and represented by "..." in the program.

[5] Note that the domain values of an enumeration may be assigned to integer values as e.g. done for the attribute *Band.Type* in Line 12.

**Listing 4.1** Feature model of an event organized by an event agency (cut-out)

```
1  root feature Event {
2    enum Discount in {Gold = 8, Staff = 3, None = 0};
3    feature Carpet {
4        int Length;
5        int Breadth in [50..300];
6        enum Color in {Red, Gold, Black, Blue};
7        bool SlipResistance;
8        int Cost is ...
9    }
10   feature Band : optional {
11       requires Stage;
12       enum Type in {Classic = 1000, Blues=700, Rock=900};
13   }
14   feature Stage : optional { ... }
15   constraint BluesOnBlueCarpet
16       Band is selected and Band.Type = Blues ->
17           Carpet.Color = Blue
18   }
19   feature ColoredBalloons {
20     int Amount in {500, 1000, 2500, 5000, 10000};
21     feature PriceReduction {
22        int Coupon ifIn:  is 1000
23                   ifOut: is 0;
24     }
25     int Cost is Amount * 3 - PriceReduction.Coupon;
26   }
27   int OverallCost is Carpet.Cost + Band.Type + ... +
28       (ColoredBalloons.Cost + ...) * (100-Discount)/100;
29 }
```

and deleted, resp.) of the feature (here *PriceReduction*). Furthermore, it is possible to define *constraints* on attributes and features, also accessing the configuration state of a feature as shown in Line 15. This constraint makes sure that the Blues-band plays in an adequate ambiance.

The *transformation* into a Choco CSP is straightforward, for details see [19]. In general, our transformation is similar to these of [15,17]. Differences come from the fact that the transformation target of these approaches are CLP languages and they aim at feature model analysis in contrast to interactive configuration, as does *FdConfig*. We show an example of the generated CSP in a mathematical notation and leave out the Choco constraint syntax for reasons of space limitations.

*Example 5.* The following CSP is generated from Listing 4.1 (where $CBal$ stands for *ColoredBalloons*, $PRed$ for *PriceReduction*, and $SRes$ for *SlipResistance* resp.). Note that we do not enumerate the set of variables $X$ explicitly and give the domains $D$ by means of element constraints $C_{Domains}$.

$$
\begin{aligned}
CSP &= C_{Domains} \land C \text{ with} \\
C_{Domains} &= Event, Carpet, Band, Stage, CBal \in \{False, True\} \land \\
&\quad CBal.PRed, Carpet.SRes \in \{False, True\} \land \\
&\quad Discount \in \{0, 3, 8\} \land Band.Type \in \{700, 900, 1000\} \land \\
&\quad Carpet.Length \in [-2^{31}, 2^{31} - 1] \land Carpet.Breath \in [50, 300] \land \\
&\quad Carpet.Color \in [0, 3] \land Carpet.Cost = ... \land \\
&\quad CBal.Amount \in \{500, 1000, 2500, 5000, 10000\} \land \\
&\quad CBal.Cost \in [-2^{31}, 2^{31} - 1] \land \\
&\quad CBal.PRed.Coupon \in [-2^{31}, 2^{31} - 1] \land \\
&\quad OverallCost \in [-2^{31}, 2^{31} - 1] \text{ and}
\end{aligned}
$$

$$
\begin{aligned}
C = {} & (Carpet \lor Band \lor Stage \lor CBal \to Event) \land \\
& (CBal.PRed \to CBal) \land (Band \to Stage) \land \\
& ((Band \land Band.Type = 700) \to Carpet.Color = 3) \land \\
& (CBal.PRed \to CBal.PRed.Coupon = 1000) \land \\
& (\neg CBal.PRed \to CBal.PRed.Coupon = 0) \land \\
& (CBal.Cost = CBal.Amount * 3 - CBal.PRed.Coupon) \land \ldots
\end{aligned}
$$

### 4.2 Domain Analysis

In *FdFeatures* the specification of an attribute's base domain is optional. If no domain is given by the user, as e.g. for *Carpet.Length* or *ColoredBalloons.Cost* in Listing 4.1, it is set by default to the maximal possible domain of the corresponding attribute type. For example, for integer attributes the maximal domain is $[-2^{31}, 2^{31} - 1]$ which we denote by $MAXDOM$ in the following.

When the Choco solver computes the valid domains of the CSP in the second phase of our approach (cf. Sect. 4.3), this may become time consuming. The solver must establish global consistency. Thus, up to $4.3 * 10^9$ values must be checked for every attribute (or its corresponding variable, resp.) with $MAXDOM$. Of course, we cannot require the user to specify attribute domains just big enough to contain all solutions, in particular, because a manual estimation of the base domain can be very difficult for complex feature models. Thus, we apply an automatic pre-analysis to the feature model which is merged with the CSP generated from the model.

Our *domain analysis* aims at an approximate yet quick pre-calculation of the base domains of variables using knowledge about the feature model's structure. We only consider integer attributes, as enumerated attributes will in general have small domains. The analysis is based on interval arithmetics [16] which allow a fast approximation of the variable's minimum and maximum values by calculating with intervals instead of single domain values.

The domain $DOM_{FM}$ of an attribute in *FdFeatures* can be specified directly by giving a single value or a set or interval, resp. of values. Additionally, it is possible to specify particular sub-domains depending on the configuration state, i.e. $IN_{FM}$ and $OUT_{FM}$ in case the attribute is selected or deleted, resp. Furthermore, arithmetic expressions can be used to specify the domain or sub-domains. We determine $DOM_{FM}$, $IN_{FM}$, and $OUT_{FM}$ in form of intervals from the attribute expressions, where enumarations are handled as intervals, too.

Starting from these domains, we calculate the narrowed base domain $BASEDOM$, and new sub-domains $IN$ and $OUT$ as follows (where we take arithmetic expressions into consideration):

$$BASEDOM = (IN_{FM} \cup OUT_{FM}) \cap DOM_{FM} \tag{1}$$
$$IN = BASEDOM \cap IN_{FM} \tag{2}$$
$$OUT = BASEDOM \cap OUT_{FM} \tag{3}$$

The intervals for the incorporated arithmetic expressions are determined by traversing their formula tree. The leafs are either elementary expressions or references to other attributes, in case of which the domain of the referenced attribute must be calculated first. The analysis of cyclic formulae is interrupted and $MAXDOM$ is used instead, leaving domain narrowing to the CHOCO solver, which uses accurate but time consuming consistency techniques.

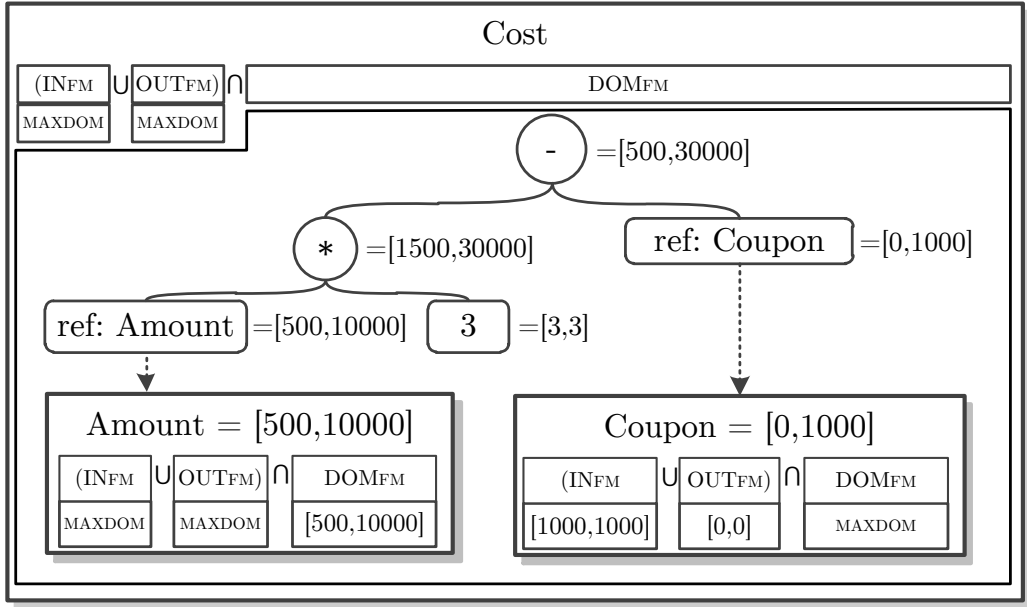

**Fig. 2.** Domain analysis of $BASEDOM$ of the attribute *ColoredBalloons.Cost*

*Example 6.* Consider the pre-calculation of the base domain $BASEDOM$ of the attribute *ColoredBalloons.Cost* (Line 25 of Listing 4.1). Figure 2 illustrates the calculation.

For the attribute under consideration, only the set $DOM_{FM}$ is specified by means of an arithmetic expression, while $IN_{FM}$ and $OUT_{FM}$ both default to $MAXDOM$. During the analysis, the formula tree of the arithmetic expression is traversed. Dashed arrows depict the domain analysis of a referenced attribute which is shown in it's own box.

In the beginning the analysis moves to the first leaf: a reference to the attribute *Amount*. The determination of the base sets is trivial as only $DOM_{FM}$ is defined as an enumeration of integers which yields an interval [500, 10000] using Equation 1. The right operand of the multiplication is a constant value, which is turned into the point interval [3, 3] resulting in the intermediate result $[500, 10000] * [3, 3] = [1500, 30000]$. The analysis of the attribute *Coupon* yields $BASEDOM = [0, 1000]$ from $IN_{FM} = [1000, 1000]$, $OUT_{FM} = [0, 0]$ and $DOM_{FM} = MAXDOM$ (again using Equation 1). Finally, the analysis returns to the root attribute *ColoredBalloons.Cost* and performs the subtraction with result $BASEDOM = [1500, 30000] - [0, 1000] = [500, 30000]$.

From the $BASEDOM$ intervals of the attributes the respective sub-domains $IN$ and $OUT$ can be inferred by means of the Equations 2 and 3 (not shown in the figure).

*Example 7.* (continuation of Examples 5 and 6) The domain analysis yields the following domain constraints as an update on the generated CSP of our event feature model.

$$C'_{Domains} = \ldots \wedge$$
$$\quad\quad CBal.Cost \in [500, 30000] \wedge$$
$$\quad\quad CBal.PRed.Coupon \in \{0, 1000\} \wedge \ldots$$

Note, that for computed intervals we finally build intersections in case the domain was initially given by enumerations or single values. This yields the two-element set for $CBal.PRed.Coupon$.

### 4.3 The Configuration Phase

The second phase of our approach, i.e. the configuration phase, starts with the initialization of *FdConfig* before the user can start with the interactive configuration process.

*Model pre-processing.* The solver reads the CSP-model and performs a feasibility check (e.g. by finding the first solution). If successful, the configurator computes the valid domains as initial *model consequences* that derive from the CSP. The calculation of these model consequences is performed in the same way as the user consequences are calculated later on in the interactive user configuration phase (see below). However, once the model consequences have been computed, they are immutable during the interactive configuration as they don't depend on the user decisions.

The current, global consistent state of the solver is recorded. To this *ground level state* the solver can be reset when, after a retraction of user constraints, a re-computation of the valid domains becomes necessary.

*User configuration.* The user starts a configuration step by executing a configuration action. This is either a configuration decision, i.e. limiting the domain of a feature- or attribute variable which manifests as a user constraint or the retraction of a decision made earlier. In this case the corresponding user constraint is removed from the constraint system. User decisions are posted by *FdConfig* to the solver as user constraints.

Now, the solver is activated to establish global consistency and to find all solutions of the constraint system. These are evaluated to derive the valid domains. Since the valid domains define the configuration options available to the user in the next configuration step, the constraint system always remains feasible after a user decision.

After the user consequences have been computed, the user interface is updated accordingly and the user can perform the next configuration action.

In the usual modus operandi for FD solvers, a CSP is once declared and then read by the solver which computes and returns solutions. In contrast, for interactive configuration we need to re-calculate sets of solutions again and again because a sub-set of the constraints (the user constraints) keeps changing over time as a result of the user making configuration decisions.

As the solver maintains a heavyweight internal representation of the constraint system and reading the CSP-model as well as establishing consistency are time consuming, the option of re-creating the solver for every user decision is inapplicable. Therefore we control the solver from outside by utilizing its backtracking infrastructure and reset the solver into the aforementioned ground level state in case a user decision has been retracted.

### 4.4 Improving the User Experience by Multithreading

When computing the valid domains of the variables, the constraint solver must establish global consistency, and thus, potentially find all solutions of the CSP. This calculation may be time consuming depending on the size and complexity of the feature model (and the CSP it was transformed into, resp.). Furthermore, the GUI would not be updated or process user input during this calculation. The program would appear to be frozen.

Therefore we introduced multithreading with the solver running in a background thread, thus, allowing the GUI to be updated and accept user input during a long running computation. However, as the user would still have to wait for the calculation to complete before he can enter another configuration decision, the multithreading structure has been extended as follows:

The elements of the valid domains are collected gradually with the computation of the set of solutions still in progress. Whenever new elements have been found, they are immediately displayed in the GUI and made available for configuration decisions. Elements, that did not yet occur in a solution, are greyed out and disabled for user decisions. If the user makes a decision, the background calculation is interrupted and restarted with the changed set of user decisions.

In the sequential model the valid domains were calculated in one go and then evaluated to generate consequence decisions if necessary. If, for example, the valid domain of a feature $A$ was found to be $D_{vd,A} = \{true\}$ this resulted in a consequence decision forcing the feature to be *selected*[6].

With multithreading we have to re-evaluate the valid domains whenever new elements are found during the calculation process. This results in changing consequence decisions while the computation has not finished. For example, the valid domain of feature $A$ can become $D_{vd,A} = \{true\}$ during the computation process at first, creating the consequence decision that $A$ must be selected. However, as the result of new solutions the valid domain might later become $D_{vd,A} = \{true, false\}$, thus making the consequence decision disappear again. Attributes are handled similarly, as single value domains (interpreted as consequence decisions to select this particular value) may become multi-value domains later on. The GUI flags these consequence decisions as *incomplete*, so the user can see that further configuration options might become available. On the completion of the computation process, this flag is removed.

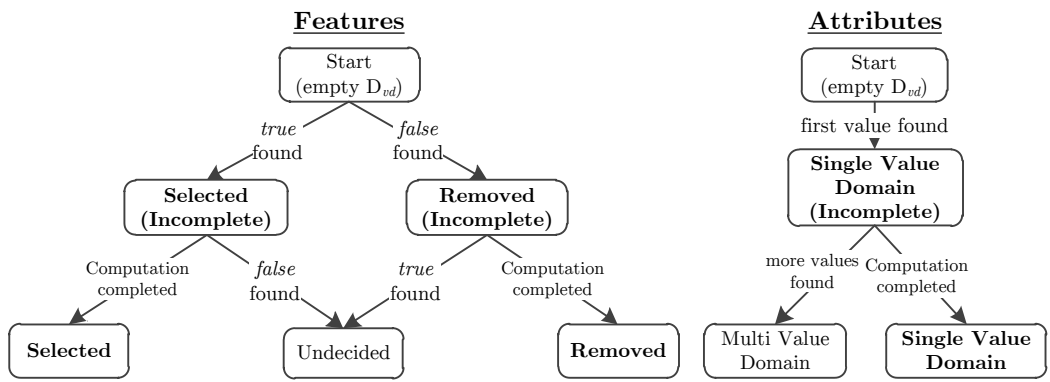

**Fig. 3.** Intermediate states of valid domains for features and attributes with multithreading

---

[6] Likewise $D_{vd,A} = \{false\}$ results in a *removed* feature and $D_{vd} = \{true, false\}$ in the *undecided* state, where the user can decide.

Figure 3 illustrates the different states for feature and attribute domains, resulting from the multithreading approach. Consequence decisions are drawn in bold typeface. Furthermore Fig. 4 shows a screenshot of the *FdConfig* tool during a long running calculation of the valid domains. Incomplete consequence decisions are visible, i.e. for the attribute *ColoredBalloons.Cost*, whose valid domain has exactly one element (14990) at the moment. The other elements were either eliminated by the user or have not yet occurred in a solution (displayed in grey).

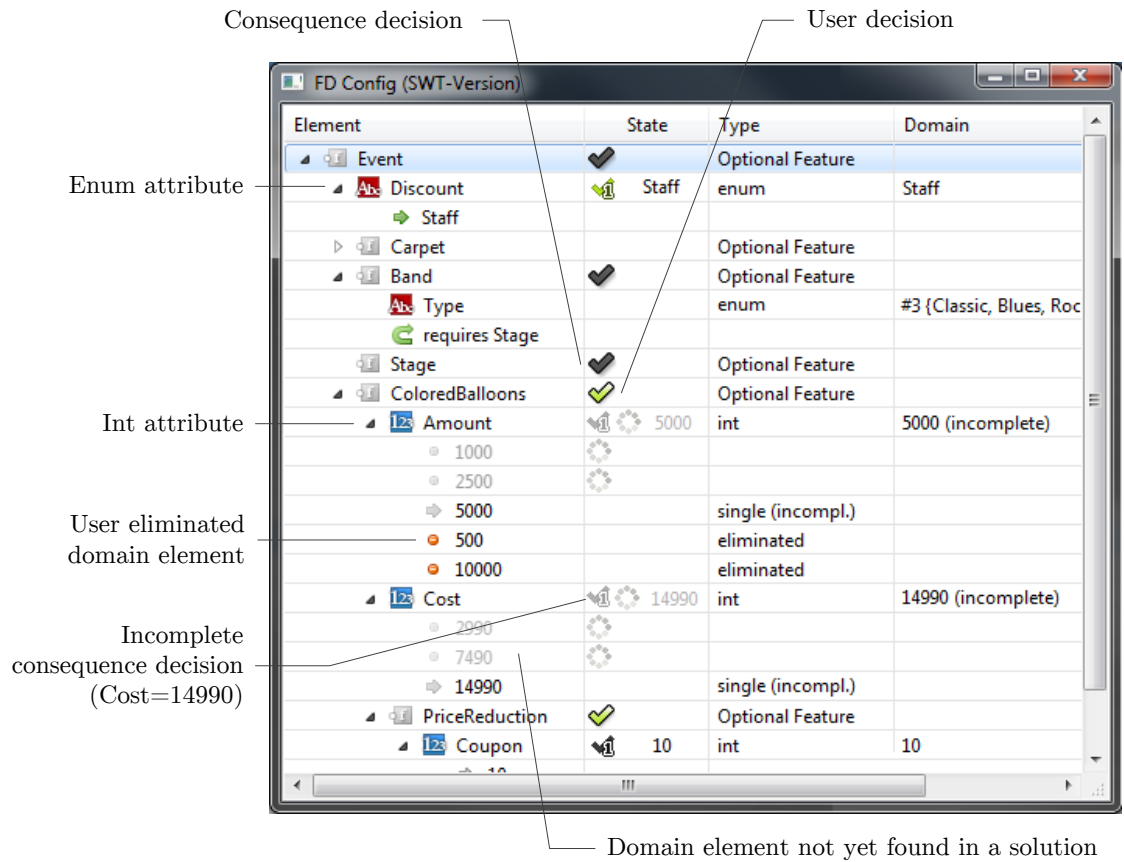

**Fig. 4.** Screenshot of *FdConfig* during a CVD calculation

First experiments show that this multithreading approach leads to a smoother, more fluent user experience when performing product configuration. Since reaching the goal of calculating the valid domains in under 250 msec[7] is currently not realistic with the underlying solvers, this enhancement is a good compromise as configuration options will become available very quickly.

## 5   Conclusion and Future Work

In this paper we discuss an approach on interactive product configuration based on constraint techniques, which was implemented in our configurator tool *FdConfig*. We gave an overview of the product configuration domain, feature models, and constraint programming in this context and introduced our approach.

---

[7] A response time of about this duration is considered desirable, as this still gives the user the impression to work in real time [7].

In *FdConfig* we employ a finite domain constraint solver that enables us to deal with integer attributes and arithmetic constraints in extended feature models. These constraints are usually not supported in traditional approaches (e.g. SAT, BDDs) or only in restricted forms. However, this enhanced expressiveness comes at the cost of performance penalties. We deal with this by applying a preliminary domain analysis in order to relief the solver of unnecessary computation time for establishing consistency. Furthermore, we use a multithreading approach to enhance the user experience. This allows the user to continue configuring in a limited way, even if the overall computation has not yet finished.

*Future work* will include the further development of the multithreading approach. We plan to incorporate multiple solvers that might use different computation strategies. For example, the feature model element with the current GUI focus could be taken into account. This focus-based computation strategy could additionally improve user friendliness: Domain elements, that the user might want to configure most likely, would become available more quickly for configuration decisions.

Also a more subtle handling of the non-chronological retraction of constraints promises improvement but needs further investigation.

In order to improve the overall performance we consider adding support for compilation-based approaches (i.e. BDDs). These could be integrated with the solver in the form of custom constraints to speed up the search. If a pre-compiled version of a feature model is available, the implementation of these constraints could access the BDD. Otherwise the regular solution methods would be applied.

Transformation-based optimizations should be investigated, too. E.g. [13] use a clustering optimization to reduce the number of constraint-variables and constraints. Using feature models ([13] directly use constraints) may support or even inherently realize a form of clustering.

Another optimization is presented in [17]. The authors discuss the improvement of efficiency when solving CSPs as transformation results due to a reformulation of particular boolean constraints into arithmetic constraints. While this representation is available in our approach too, the examination of similar optimizations may be worth considering in the future. In the approach of [17] the structure of feature models is not preserved. This holds optimization potential as well, but must be done sensitive to retain a mapping to the feature model to allow an interactive configuration process as needed in our approach.